\documentclass[10pt,twocolumn,letterpaper]{article}

\usepackage[pagenumbers]{cvpr}      
\usepackage{multirow}
\usepackage[normalem]{ulem}
\useunder{\uline}{\ul}{}
 \usepackage{graphicx} 
 \usepackage{colortbl}

\definecolor{cvprblue}{rgb}{0.21,0.49,0.74}
\usepackage[pagebackref,breaklinks,colorlinks,allcolors=cvprblue]{hyperref}

\def\paperID{11536} 
\def\confName{CVPR}
\def\confYear{2025}
\usepackage{cinzel}
\def\benchmark{\textcinzel{\textit{UIX}}plore}
\def\modelname{Auto-Explorer}

\usepackage[dvipsnames]{xcolor}

\title{AUTO-Explorer: Automated Data Collection for GUI Agent}

\author{
Xiangwu Guo \ \ \ \ \ \      Difei Gao \ \ \ \ \ \   Mike Zheng Shou\thanks{Corresponding author.} \\
Show Lab, National University of Singapore
}

\begin{document}
\maketitle
\begingroup
\renewcommand\thefootnote{$*$}

\endgroup

\begin{abstract}
Recent advancements in GUI agents have significantly expanded their ability to interpret natural language commands to manage software interfaces. However, acquiring GUI data remains a significant challenge. Existing methods often involve designing automated agents that browse URLs from the Common Crawl, using webpage HTML to collect screenshots and corresponding annotations, including the names and bounding boxes of UI elements. However, this method is difficult to apply to desktop software or some newly launched websites not included in the Common Crawl. While we expect the model to possess strong generalization capabilities to handle this, it is still crucial for personalized scenarios that require rapid and perfect adaptation to new software or websites. To address this, we propose an automated data collection method with minimal annotation costs, named \modelname. It incorporates a simple yet effective exploration mechanism that autonomously parses and explores GUI environments, gathering data efficiently. Additionally, to assess the quality of exploration, we have developed the \benchmark~benchmark. This benchmark creates environments for explorer agents to discover and save software states. Using the data gathered, we fine-tune a multimodal large language model (MLLM) and establish a GUI element grounding testing set to evaluate the effectiveness of the exploration strategies. Our experiments demonstrate the superior performance of \modelname, showing that our method can quickly enhance the capabilities of an MLLM in explored software.
\end{abstract}    
\section{Introduction}
\begin{figure*}[htbp]
    \centering
    \includegraphics[width=\textwidth]{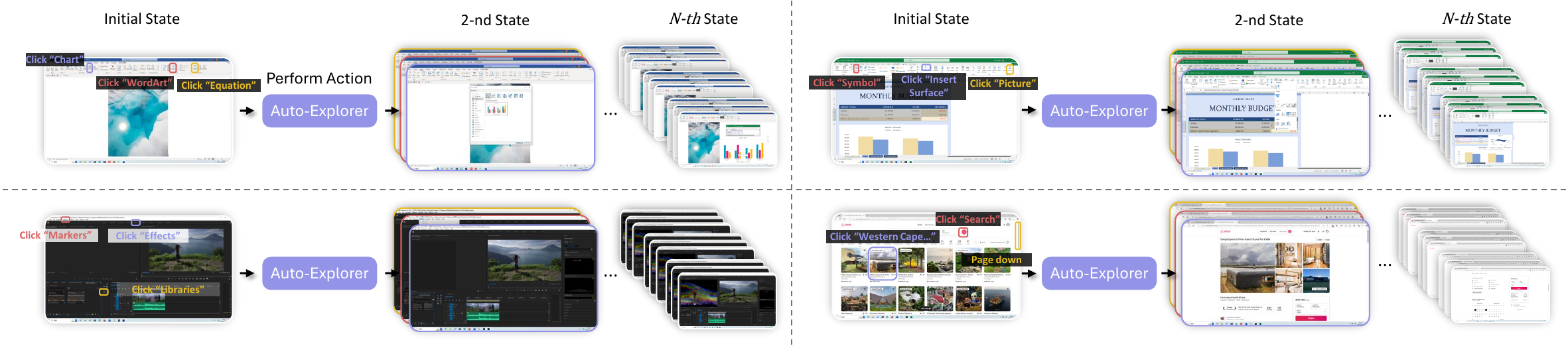}
    \caption{\textbf{\modelname} is a data collection agent capable of autonomous exploration within a given environment. It utilizes various tools such as UI Automation (UIA), Optical Character Recognition (OCR), and icon template matching algorithms to parse the content of images. The agent selects UI elements to interact with from these parsed elements, continuously exploring new environmental states.}
    \label{fig1}
\end{figure*}

GUI agents~\cite{sun2022metagui, mind2web, cogagent, assistgui,webagentplan,yao2022webshop,copilot,ferretui,guicourse,guiworld,zhang2024aitz}, which can interpret natural language commands to operate software interfaces, have recently captured significant interest due to their potential to revolutionize human-computer interaction. By bridging the gap between language and action, these agents promise to streamline complex workflows~\cite{wu2024copilot, osworld} and enhance productivity across diverse software applications. 


While many efforts have made significant progress, realizing this vision faces substantial challenges, mainly due to the scarcity of annotated training data tailored for GUI environments. Existing methods~\cite{cogagent, seeclick} typically employ an automated agent to navigate URLs from the Common Crawl, using webpage HTML to gather screenshots and corresponding annotations, including the names and bounding boxes of UI elements. For desktop applications or newly launched websites, the absence of a universal identifier like a URL complicates the process of switching between different application states, making it difficult to collect diverse screenshots. On the other hand, websites can easily acquire annotations for GUI elements. In contrast, for desktop applications, while some software supports annotations through UI Automation (UIA), many others do not due to compatibility issues. These issues often stem from older software architectures that do not integrate modern accessibility features or from applications developed with custom, non-standard user interface components that UIA cannot readily identify and interact with.

To address the data collection issue, we introduce an automatic GUI explorer, \modelname, which can automatically perform actions over the software or website to discover the states and save the results, as shown in Figure~\ref{fig1}. It is done by two core modules: a GUI parser that can detect the UI elements and an explorer module to perform the actions. The GUI parser is able to use HTML, and UIA, but more importantly, it can also use OCR for text element detection and template matching for icon detection when HTML and UIA fail. And Explorer module is a simple yet effective exploring strategy. The basic idea is that, in the initial state, the model will randomly click on a button that has not been clicked before. After each action, the agent compares the elements before and after the action to discover new ones. It then randomly samples an action from the newly discovered elements to execute. If no new elements appear after executing a particular action, it stops.

As data collection has emerged as an essential component for training GUI agents, various studies have proposed their own mechanisms for gathering interface data. To establish a more consistent and reliable standard, we developed\benchmark~benchmark specifically designed to assess data exploration quality. Specifically, the benchmark provides 10 initial environments spanning different software and website categories — each representing an opened project file (if needed) for a software application or the main page of a website. The agent is tasked with performing an exploration task for each environment, where the goal is to execute a predefined number of GUI actions (e.g., click, drag, scroll) to explore various states within each environment. The agent stores all screenshots and parsed results of these discovered states, forming a dataset. The quality of the gathered samples is evaluated based on the effectiveness of the data in fine-tuning MLLM. In addition, we construct a GUI element grounding testing set consisting of \textbf{4,800} GUI element grounding samples with diverse query types for testing their performance. This assessment offers insights into the exploration method's capability to capture a broad range of interactions and valuable data across diverse software and web environments.

Through extensive experiments, we validate the effectiveness and adaptability of our approach across various exploration strategies. By comparing our method to alternative exploration techniques, we demonstrate significant improvements in both the efficiency and coverage of UI element detection within software environments. Our model not only achieves a higher rate of unique action and screenshot collection but also improves the understanding of GUI elements.

In summary, the main contributions of our paper are threefold:
\begin{itemize} 
\item We propose an automatic data collection method to effectively gather GUI data for novel applications or websites.
\item We construct a benchmark to evaluate the explorer agent's capability in data collection. 
\item We conduct thorough experiments to analyze existing data collection strategies and our proposed \modelname. Additionally, we provide insights into the future direction of the data explorer. 
\end{itemize}

\section{Related Work}

\textbf{GUI Agent.}
The field of Graphical User Interface (GUI) agents~\cite{miniwob++,assistgui,gpt4vsom,osworld} has seen significant advancements, particularly with the integration of large language models (LLMs) and multimodal language models (MLLMs)~\cite{openai2024gpt4o, touvron2023llama, zhu2024llava} capable of generating actions~\citep{weng2023prompt,sumers2023cognitive, hong2024metagpt, gao2023assistgpt}. Early efforts~\cite{gpt4vsom,assistgui} in this domain focused on training-free agents, which leveraged LLMs' reasoning abilities~\cite{react, yao2023tree} to generate actions from a UI representation similar to HTML. These approaches demonstrated the potential of achieving good performance without requiring any additional training. However, the tools for extracting GUI elements typically faced significant limitations, often supporting only web-based applications or a limited set of software environments. In response to these challenges, subsequent research~\cite{cogagent,seeclick,fuyu-8b,ferretui,baechler2024screenai} introduced purely visual solutions. These methods developed MLLMs~\cite{wang2023visionllm, bai2023qwen, chen2023minigpt} to handle UI elements directly from raw visual input, enabling action generation without relying on structured representations. Further works~\cite{uground} combined the long-term planning abilities of LLMs with the grounding power of MLLMs, resulting in agents capable of more complex reasoning tasks and a broader understanding of various GUIs. Such approaches have shown promise in scaling the functionality of GUI agents to a wider range of environments, from web interfaces to native desktop applications.

\noindent \textbf{GUI Dataset and Collection Strategy.}
Unlike traditional multimodal datasets, which often draw from abundant natural image data available on the internet, GUI data is inherently scarce. Many efforts have introduced innovative methods to gather this data. Some approaches, such as those proposed by CogAgent and SeeClick~\cite{cogagent, seeclick}, employ automated scripts to browse web pages extracted from Common Crawl, using tools like Playwright to render each page and capture its elements along with their bounding boxes. However, this method is limited to web-based data collection. For desktop applications, many are not able to get element coordinates by using system tools like UI Automation (UIA). In addition, unlike web pages, there is no universal identifier (such as a URL) to quickly switch between different application states. To address these challenges, some projects~\cite{agentstudio, guicourse} have developed custom annotation tools, enabling users to manually label UI elements' bounding boxes. Yet, this approach remains challenging to scale to larger datasets. 

In this paper, we introduce a new method for data collection. Our approach involves a simple template-based icon detection mechanism that requires only icon annotations to interpret screenshots. Additionally, we propose an automated exploration strategy that efficiently navigates through various application states, capturing diverse screenshots to enrich the dataset.

\noindent \textbf{GUI Related Benchmark.}
Most existing GUI-related benchmarks~\cite{webarena, mind2web, miniwob++, assistgui, aitw} focus on evaluating the execution capabilities of various GUI tasks, such as shopping~\cite{yao2022webshop}, web navigation~\cite{androidworld}, and productivity tasks~\cite{osworld,assistgui}. Other works~\cite{seeclick,omniact} assess the grounding capabilities of GUI elements or test the understanding of atomic GUI actions~\cite{act2cap}. Additionally, more recently work~\cite{videogui} has evaluated comprehension of instructional videos related to GUI tasks. In this paper, we propose a benchmark for a unique type of agent for data collection, evaluating their ability to autonomously explore and reach different states for a specific software or website. This benchmark serves as a measure of an agent's effectiveness as a data collector in the new domain.

\section{\texorpdfstring{\benchmark~Benchmark}{Benchmark Benchmark}}

\begin{figure}[tbp]
    \centering
    \includegraphics[width=\columnwidth]{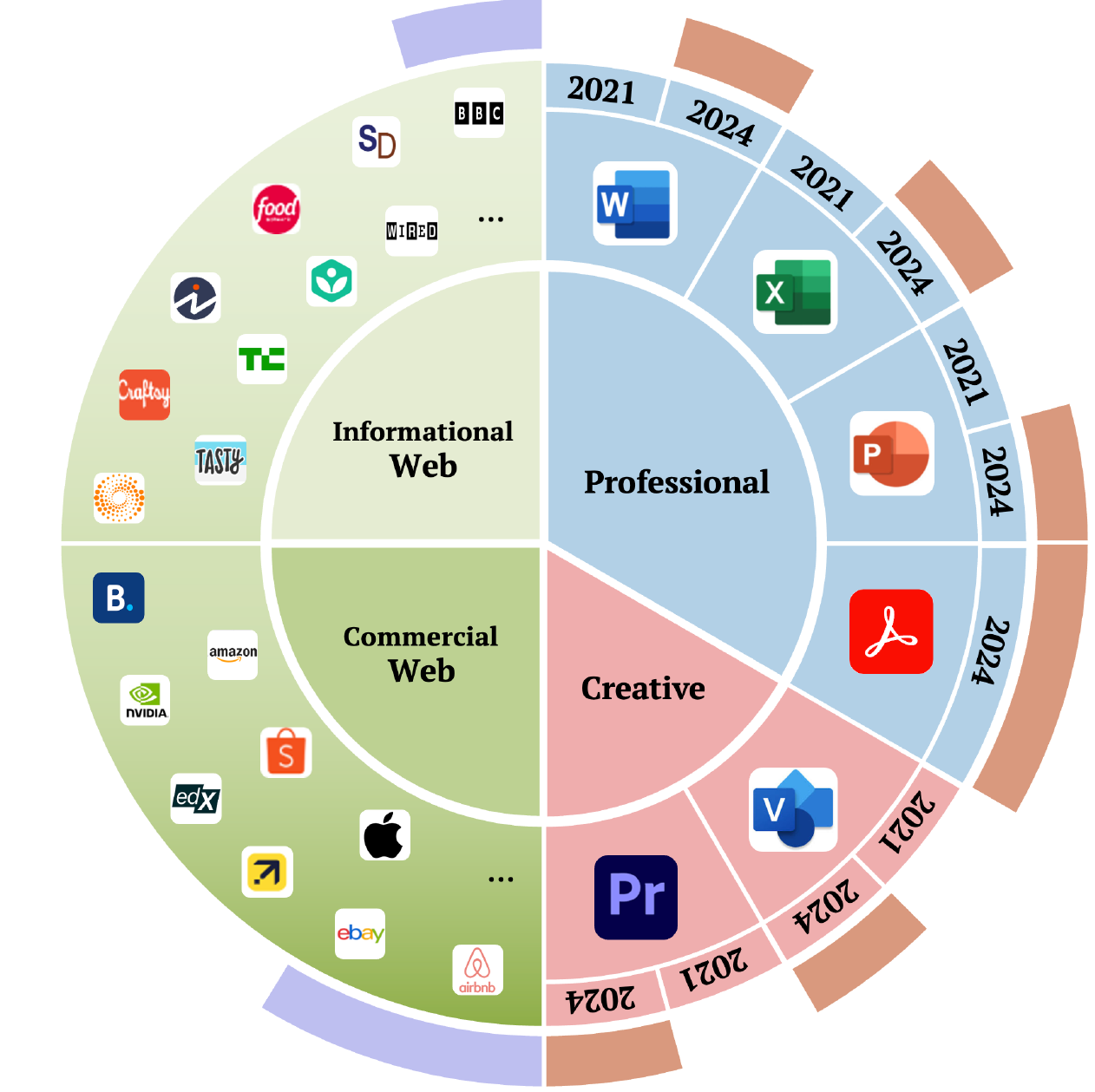}
    \caption{This diagram categorizes software and websites used in our benchmark, with the outer ring's color coding indicating usage: {\color{brown}\textbf{light brown}} for software or websites used in both exploration environments and GUI grounding testing set, {\color{violet}\textbf{purple}}  for exploration-only ones, and {\color{gray}\textbf{uncolored}} for the testing set.}
    \label{fig:software}
\end{figure}

\subsection{Task Formulation}
The objective of this benchmark is to assess an exploration agent's ability to collect data from novel software applications and websites, measuring both the diversity and effectiveness of the collected data.

Specifically, the benchmark provides 10 initial environments—each representing an opened project file (if needed) for a software application or the main page of a website. The agent is tasked with performing an exploration task for each environment, where the goal is to execute a predetermined number of GUI actions (e.g., click, drag, scroll) to explore various states within each environment. The agent stores all screenshots and parsed results of these discovered states, forming a dataset.

The quality of the gathered samples is evaluated based on the number of unique actions explored and the effectiveness of the data in fine-tuning MLLM. This assessment offers insights into the exploration method's capability to capture a broad range of interactions and valuable data across diverse software and web environments.





\subsection{Environments Setup}
All environments in\benchmark~are built on the Windows OS. In each environment, a software application with a specific version and an associated project file (if needed) or the main page of a website will be open, as the initial state. As shown in Figure~\ref{fig:software}, We provide environments across diverse applications and websites, categorized into four main groups:

\noindent \textbf{Software Applications:}
\begin{itemize}
\item \textbf{Productive:} This category includes PowerPoint, Word, Acrobat, and Excel, which are among the most widely used office software globally. Each application is explored across multiple versions and project files, providing diverse data sources.
\item \textbf{Creative:} This includes complex software used by professionals, such as Visio and Adobe Premiere. Exploration is conducted on various versions and project files to capture a wide range of design functionalities.][o]
\end{itemize}

\noindent 
\textbf{Web Sources:}
\begin{itemize}
\item \textbf{Commercial:} This category includes websites offering transactions or services, such as those related to travel, shopping, and education. These sites provide a variety of interactive elements to explore.
\item \textbf{Informational:} This includes non-profit websites that offer information or assistance, such as news outlets, DIY-sharing platforms, and food-sharing sites. These sites offer a rich array of informational content.
\end{itemize}

\subsection{Exploration Task}
Given the initial environment, the agent is required to perform 500 diverse GUI actions, including clicks, drags, and other possible interactions. The agent needs to save all screenshots throughout the exploration process and must design a custom screenshot parser to capture the corresponding parsed data, including UI element names and bounding boxes. These data will be stored to form a training dataset.

\subsection{Evaluation}

To evaluate the effectiveness of the exploration method used in \benchmark, we use collected data to fine-tune the same multimodal model, Qwen2-VL-2B~\cite{qwen2vl} with the same training strategy. Then, we will test its performance of GUI element grounding on a human-labeled testing set.

\textbf{GUI Element Grounding.}
The collected data will be used to train the GUI element grounding task. Specifically, given a natural language query that describes a UI element, the fine-tuned model is required to generate the corresponding bounding box.

\begin{table}[]
\centering
\resizebox{\columnwidth}{!}{\Large
\begin{tabular}{p{1.5cm} p{12cm}}
Type                            & Instruction \\
\midrule
{\color[HTML]{009901} Name}     & Find \color[HTML]{009901}“Star these file icon” \\
{\color[HTML]{F56B00} Shape}    & Find the element which has the following description: \color[HTML]{F56B00} A gray star with an empty center \\
{\color[HTML]{6665CD} Function} & Find the element which has the following function: \color[HTML]{6665CD} Marks the file as a favorite or important for easy access \\
{\color[HTML]{00D2CB} R.E}      & Find \color[HTML]{009901}Star these file icon\color[HTML]{000000}. The surrounding information is: \color[HTML]{00D2CB} To the right of “Save files icon” and to the left of “Save these file to Adobe Cloud icon” in the toolbar at the top of the page \\
\end{tabular}%
}
\caption{\textbf{Types of Instructional Data} generated for model fine-tuning.}
\label{tab:instruct}
\end{table}

\textbf{Model Fine-tuning.} 
Given screenshots and UI element annotations, including the UI element name and bounding box collected by the exploration agent, we will automatically generate instructional data. By using GPT-4o, we aim to generate diverse queries and corresponding ground-truth bounding boxes based on simple button names and their bounding boxes, as detailed in Table~\ref{tab:instruct}.
Then, these data will be used to fine-tune the Qwen2-VL-2B. Specially, These data are then used to fine-tune the Qwen2-VL-2B model. Specifically, to align with the grounding capabilities of Qwen2-VL, the model training follows its grounding format. All models in the benchmark are trained for 5 epochs with identical parameters, employing LoRA for fine-tuning.

\textbf{Testing Set Construction for GUI Element Grounding.}
Our test set consists of \textbf{4,800} GUI element grounding samples, each including a screenshot, a text query, and a ground-truth bounding box. Human annotators gathered these samples by navigating the relevant software or websites, capturing screenshots, and identifying UI elements that correspond to the queries and bounding boxes. The query types reflect those in the fine-tuning dataset and include \textbf{Name} (element's name), \textbf{Shape} (description of the element's shape), \textbf{Function} (element's function), and \textbf{Referring Expression} (the element's relationship with its surroundings, such as ownership or positional relationships). However, unlike the fine-tuning dataset, in the test set, both bounding boxes and UI element names are labeled by human annotators.

To avoid model overfitting on specific software or websites, the annotators intentionally include screenshots from different versions of software in the test set. For websites, all screenshots are from the same category but different sites. In Figure~\ref{fig:software}, we visualize which software or websites are unique to the testing set.

%



\textbf{Metrics.}
We will calculate the Intersection Over Union (IOU) between the predicted bounding box and the ground truth bounding box. If the IOU is greater than 0.3, it will be deemed correct; otherwise, it will be deemed incorrect. Ultimately, we will calculate the accuracy across all samples.

\section{\modelname}
\label{sec:methodology}

\begin{figure*}[ht]
    \centering
    \includegraphics[width=\linewidth]{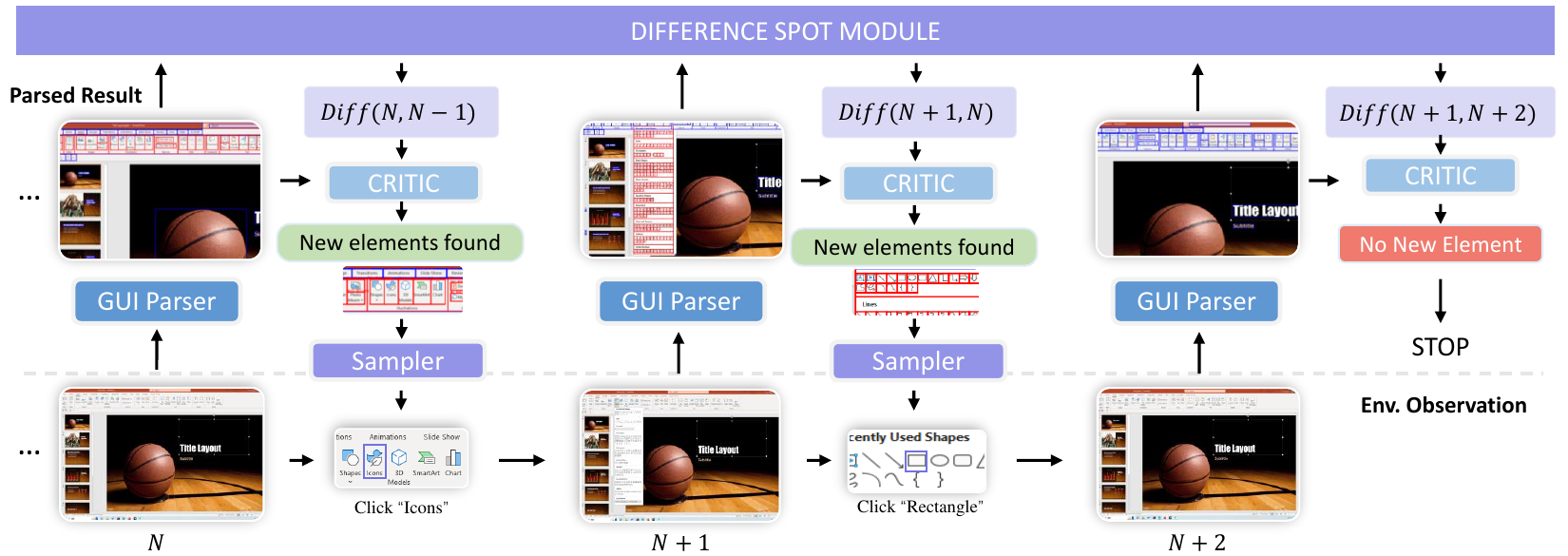}
    \caption{\textbf{Pipeline of \modelname}: \modelname~consists of two primary components: 1) GUI parser, which automatically parses UI elements from screenshots, and 2) Explore Module, tasked with determining subsequent actions to uncover new environment states. Initially, the model selects a random unclicked button. After each action, the Difference Spot Module checks if there are new elements in the UI, and then the Critic Module will choose a random action from these for execution. If no new elements emerge post-action, exploration stops.}
    \label{fig:sample}
\end{figure*}

In this section, we describe the details of \modelname, which includes two crucial components: 1) GUI parser for automatically parsing the UI elements in the screenshot, and 2) Explore Module for determining which action to perform next to discover different states of the environment.



\subsection{GUI Parser}
The GUI Parser is tasked with identifying the names and bounding boxes of all UI elements on the screen. For many websites and applications, this information can be efficiently extracted using UI Automation (UIA). However, certain applications, like Adobe Premiere Pro, do not readily expose their UI elements' details through UIA, necessitating alternative strategies for accurate detection and interaction. Specifically, for text



\textbf{Text Elements Detection.} OCR (Optical Character Recognition) tools are quite powerful for detecting text UI elements. Specifically, Google Cloud OCR~\cite{googleocr} is used to detect text UI elements when UIA is unavailable, including, all elements in Premiere Pro, Adobe Acrobat Reader, and text in the Workspace of PowerPoint.

\textbf{Icon Elements Detection.} 
For scenarios where UIA is ineffective, we introduce a template-matching method for icon detection. Initially, the icon template is obtained from the software's installation directory or manually cropped if unavailable. Once acquired, the icon template is manually labeled. Subsequently, the template-matching algorithm is employed to detect screen elements containing the template image. Due to space limitations, we elaborate more details in Supplementary.




\subsection{Explore Module}
The main function of the Explore module is to determine the next exploration steps when to stop exploring based on the \textbf{current screen} and \textbf{historical trajectory}, and how to restart exploration from the initial state.

The basic approach is that, in the initial state, the model will randomly click on a button that has not been clicked before. After each action, the agent compares the elements before and after the action to discover new ones. It then randomly samples an action from the newly discovered elements to execute. If no new elements appear after executing a particular action, it stops.

\textbf{Difference Spot Module.}
After executing each action, it inputs the parsed pre-action and post-action GUI states into the Difference Spot Module, which detects any new elements by comparing the button names. These newly identified elements are then sampled and added to the queued action list for further exploration. As illustrated, after completing an action, the Explorer selects from newly appeared elements to perform action A+1, continuing this iterative process until no new elements are detected, at which point the trajectory stops.

\textbf{Critic Module.} The Critic Module is for evaluating the significance of each interaction by analyzing the state of the GUI before and after actions. It performs three main functions:
\begin{itemize}
\item \textbf{Trajectory Termination:} It determines when to stop a trajectory. By assessing whether the Difference Spot Module has identified any new elements, the Critic Module decides whether to proceed with further actions or terminate the current trajectory. This ensures that resources are not wasted on fruitless explorations.

\item \textbf{Error State Identification} It identifies error states. Utilizing traditional image processing techniques, GPT-4o, and GUI information analysis, the module detects anomalies such as when no changes occur after an action or when error/warning dialogs appear. When such conditions are detected, the Critic Module halts the trajectory to maintain the stability of the exploration process. Importantly, these error states are recorded, capturing all preceding actions and states. This information is invaluable for developing more robust AI agents capable of handling errors effectively.

\item \textbf{Exploration Completion:} The Critic Module determines when to stop the entire exploration process. As AUTO Explorer begins its journey from the initial state, the queued action list expands with new elements identified by the Difference Spot Module. However, as exploration deepens and most trajectories yield no new elements, the queued action list starts to shrink. When it eventually becomes empty, it signifies that the current software has been thoroughly explored. At this point, the Critic Module ceases the AUTO Explorer's operations, concluding the exploration process.
\end{itemize}






\section{Experiments}

\begin{table*}[]
\resizebox{\textwidth}{!}{%
\begin{tabular}{ccccccccccccc}
\hline
 &
   &
  \multicolumn{4}{c}{\textbf{Software}} &
  \multicolumn{4}{c}{\textbf{Web}} &
  \multicolumn{3}{c}{} \\ \cmidrule(lr){3-6} \cmidrule(lr){7-10}
 &
   &
  \multicolumn{2}{c}{\textbf{Creative}} &
  \multicolumn{2}{c}{\textbf{Productive}} &
  \multicolumn{2}{c}{\textbf{Informational}} &
  \multicolumn{2}{c}{\textbf{Commercial}} &
  \multicolumn{3}{c}{\multirow{-2}{*}{\textbf{Whole}}} \\
\multirow{-3}{*}{\textbf{Model}} &
  \multirow{-3}{*}{\textbf{Method}} &
  \textbf{Icon} &
  \textbf{Text} &
  \textbf{Icon} &
  \textbf{Text} &
  \textbf{Icon} &
  \textbf{Text} &
  \textbf{Icon} &
  \textbf{Text} &
  \textbf{Icon} &
  \textbf{Text} &
  \textbf{All} \\ \hline
\rowcolor[HTML]{FBE5D6} 
\begin{tabular}[c]{@{}c@{}}Qwen2-VL-2B\end{tabular} &
  Zero-Shot &
  0.04 &
  0.08 &
  0.07 &
  0.04 &
  0.12 &
  0.08 &
  0.14 &
  0.09 &
  0.08 &
  0.07 &
  0.07 \\
\rowcolor[HTML]{FBE5D6} 
\begin{tabular}[c]{@{}c@{}}Qwen2-VL-7B\end{tabular} &
  Zero-Shot &
  0.11 &
  0.12 &
  0.12 &
  0.10 &
  0.11 &
  0.15 &
  0.09 &
  0.13 &
  0.11 &
  0.13 &
  0.12 \\
\rowcolor[HTML]{FBE5D6} 
OmniParser &
  Zero-Shot &
  0.12 &
  \textbf{0.33} &
  0.06 &
  0.40 &
  0.33 &
  0.37 &
  0.24 &
  0.52 &
  0.14 &
  0.41 &
  0.32 \\
\rowcolor[HTML]{E2F0D9} 
\cellcolor[HTML]{E2F0D9} &
  Random Walk w. OCR &
  0.07 &
  0.11 &
  0.05 &
  0.13 &
  0.14 &
  0.17 &
  0.12 &
  0.22 &
  0.08 &
  0.15 &
  0.13 \\
  
\rowcolor[HTML]{E2F0D9} 
\cellcolor[HTML]{E2F0D9} &
  Random Walk w. GUI Parser &
  0.25 &
  0.16 &
  0.27 &
  0.28 &
  0.49 &
  0.39 &
  0.52 &
  0.35 &
  0.34 &
  0.32 &
  0.33 \\
  
\rowcolor[HTML]{E2F0D9} 
\cellcolor[HTML]{E2F0D9} &
  GPT-4o w. GUI Parser &
  0.21 &
  \underline{0.16} &
  0.19 &
  0.27 &
  0.40 &
  0.33 &
  0.44 &
  0.31 &
  0.27 &
  0.29 &
  0.28 \\
\rowcolor[HTML]{E2F0D9} 
\multirow{-4}{*}{\cellcolor[HTML]{E2F0D9}\begin{tabular}[c]{@{}c@{}}Qwen2-VL-2B\end{tabular}} &
  \modelname &
  \underline{0.30} &
  0.15 &
  \underline{0.34} &
  \underline{0.40} &
  \underline{0.59} &
  \underline{0.48} &
  \underline{0.67} &
  \underline{0.50} &
  \underline{0.42} &
  \underline{0.42} &
  \underline{0.42} \\
\rowcolor[HTML]{DEEBF7} 
\begin{tabular}[c]{@{}c@{}}Qwen2-VL-2B\end{tabular} &
  \modelname Full &
  \textbf{0.42} &
  0.24 &
  \textbf{0.42} &
  \textbf{0.52} &
  \textbf{0.67} &
  \textbf{0.58} &
  \textbf{0.70} &
  \textbf{0.57} &
  \textbf{0.50} &
  \textbf{0.51} &
  \textbf{0.51} \\ \hline
\end{tabular}%
}
\caption{\textbf{Accuracy Performance (\%) Comparison of Various Models}: This table provides a detailed breakdown of accuracy for different SOTA methods (with {\color{orange}\textbf{orange background color}}) and explorer baseline (with  {\color{teal}\textbf{green background color}}), and a reference model \modelname with more exploration actions (with {\color{cyan}\textbf{blue background color}}).}
\label{tab:Table1}
\end{table*}

\subsection{Baseline and Variant of \modelname}

To demonstrate the robustness of AUTO-Explorer, we evaluated and designed several baseline exploration strategies: 

\begin{itemize}
\item \textbf{Random Walk w. OCR Parser}: This baseline parses the GUI elements with the OCR tool, and then it randomly selects one button to click from all parsed elements. 
\item \textbf{Random Walk w. GUI Parser}: This baseline parses the GUI elements with our proposed GUI parser, and then it randomly selects one button to click from all parsed elements. 
\item \textbf{GPT-4o w. GUI Parser}: This baseline utilizes our proposed GUI parser to analyze the GUI elements. It then employs GPT-4o to select an element to click from the parsed results. Specifically, the GUIParser generates the parsed GUI of the current state. This parsed result, along with the names of elements already explored, is provided as context to GPT-4o. GPT-4o then selects an element that is likely to yield more meaningful and diverse outcomes upon interaction. More details are shown in Supp.
\item \textbf{\modelname Full.} The architecture of \modelname is the same as \modelname, but performs 3.5 times of actions in the environments.

\end{itemize}


\subsection{Comparison with SOTA and Different Explorer}

In Table~\ref{tab:Table1}, we tested the performance of Qwen2-VL-2B and Qwen2-VL-7B on our\benchmark, alongside the current state-of-the-art GUI parsing model, Omniparser, as a reference. All of these works are reported with zero-shot performance. In addition, we test different baseline explorers and one variance of \modelname, \modelname-Full.

The \modelname~Full method outperforms other methods across almost all metrics in both software and web categories. For example, it achieves a score of 0.42 for Creative icons and 0.24 for Creative text in software, and even higher scores in the web category, reaching 0.57 for Commercial icons and 0.50 for Commercial text. This indicates a significant advancement in its capability to adapt and accurately identify elements across different contexts.

The zero-shot models and Random Walk w. OCR generally yield lower performance, particularly evident in the software environment where their scores rarely exceed 0.15. This could suggest limitations in these methods' ability to generalize without explicit prior training or in dynamically changing environments.

The method incorporating GPT-4o w. GUI Parser presents an improvement over the simpler methods and performs well in the Informational and Commercial categories, particularly in the web environment, which may point towards its effective parsing and processing abilities in more structured or data-intensive contexts.

Overall, the results underscore \modelname Full's superior adaptability and robust recognition capabilities, making it highly suitable for diverse applications where accurate and dynamic element recognition is crucial, such as in adaptive user interfaces or automated content management systems.



\subsection{Impact of \# of Unique Exploration Actions}

\begin{figure}[htbp]
    \centering
    \includegraphics[width=\columnwidth]{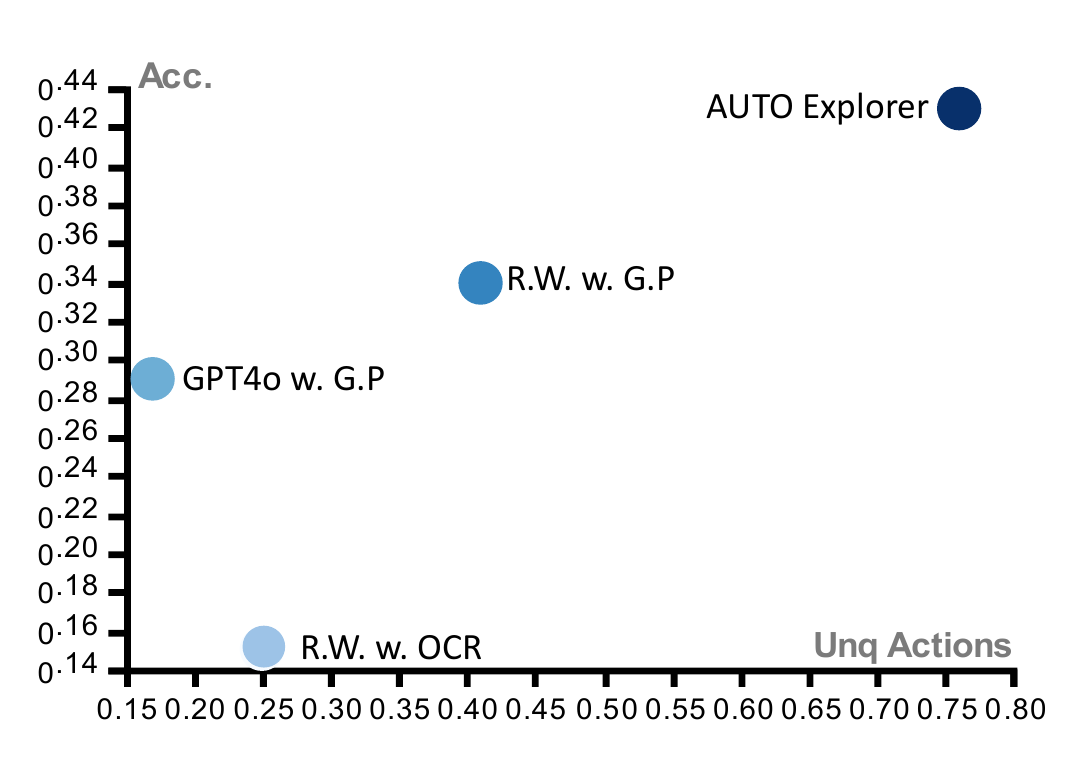}
    \caption{\textbf{Accuracy Performance} (\%) on \benchmark~against unique actions rate.}
    \label{figUni}
\end{figure}

In this section, we hope to investigate the reasons why different explorer algorithms result in poor data collection quantities. Our main hypothesis is that the relatively poor exploration strategies fail to effectively uncover new model states, leading to a high number of repeated actions. Therefore, we have analyzed the number of unique actions in baseline methods by checking for duplicates in the names of the UI elements they clicked and examining the correlation with accuracy, as shown in Figure~\ref{figUni}. It can be observed that there is a clear correlation between the number of unique actions and the final results, with Auto-Explorer producing the most unique actions. Specifically, \modelname~contains 1.85 times of example compared with Random Walk w. GUI Parser.

\subsection{Generalization Ability}
In constructing our GUI grounding testing set of \benchmark, some of the software screenshots were from environments where the collected data and test data versions were identical, while other parts consisted of different software versions. For websites, all data were of the same type but from entirely new sites. Table~\ref{tab:seen} tested the models on seen and unseen data types to evaluate the models' generalization performance. 
The results indicate that the models indeed perform significantly better on data from domains they have previously encountered. However, as the quality of the collected data improves, the models' generalization performance also markedly enhances, demonstrating rapid improvement in unseen data.

Auto-Explorer shows the highest performance across all categories, particularly excelling in unseen contexts both for software and web, with scores up to 0.64 for web icons and 0.49 for web text. This suggests Auto-Explorer's superior adaptability and robustness in handling new, variable environments compared to the other methods.

\begin{figure}[tbp]
    \centering
    \includegraphics[width=\columnwidth]{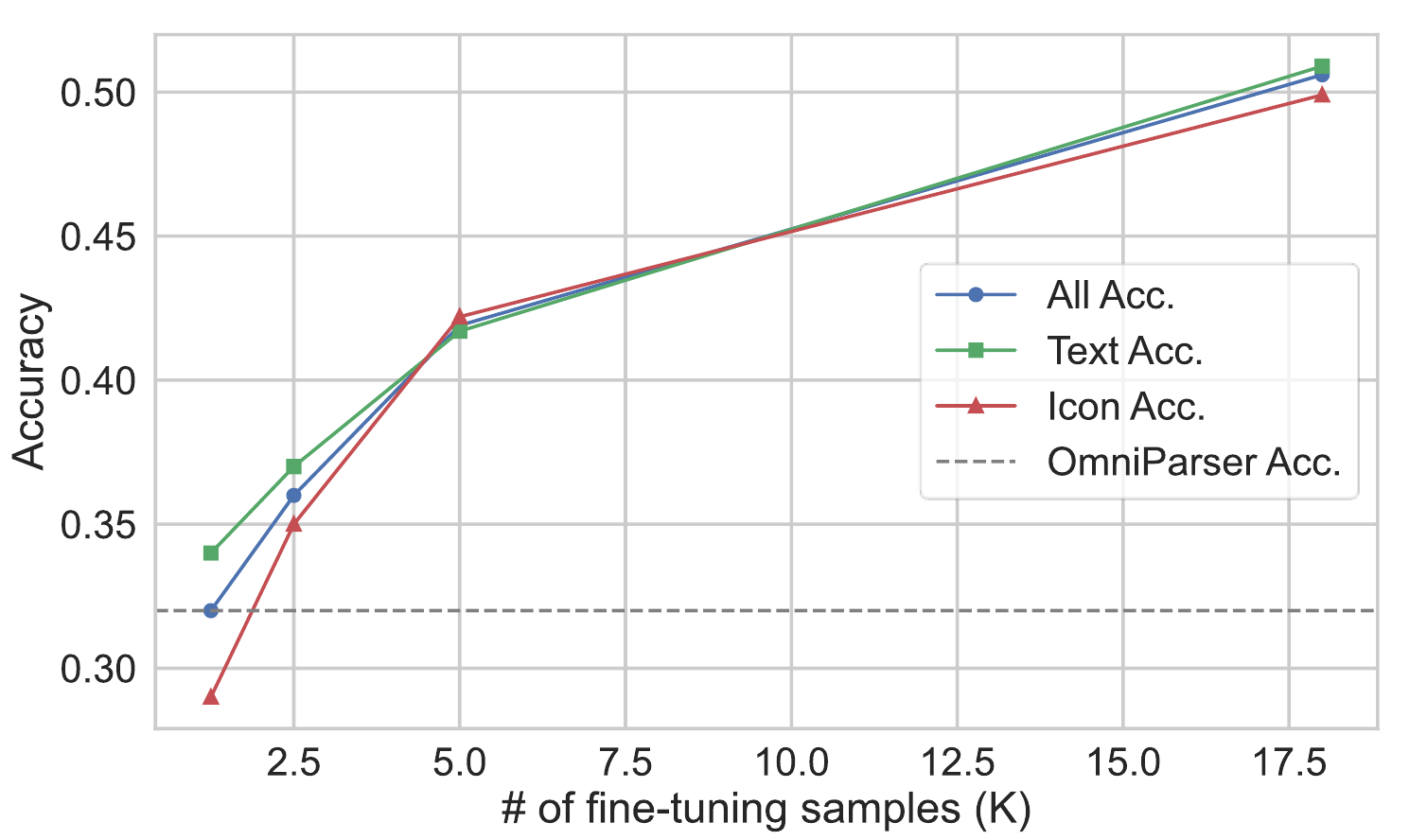}
    \caption{\textbf{Impact of Fine-Tuning Sample Size on Model Accuracy}: This graph displays the progression of accuracy improvements for All Accuracy, Text Accuracy, Icon Accuracy. The gray line represents the results of OmniParser.}
    \label{fig:data_volume}
\end{figure}

\subsection{Performance of Different Query Types}

The \benchmark~includes a variety of query types, not only grounding using button names but also incorporating functional descriptions, shape descriptions of UI elements, and referring expressions. We show the performance of the baselines and the \modelname~in a Table~\ref{tab:query}. The variation in performance across different query types indicates that the complexity and specificity of the query significantly influence the efficacy of the tested methods, with simpler visual cues like button names generally yielding better results than more abstract queries like shape descriptions. This may be because referring expressions often contain an excess of information, while button names may align more easily visually.

The \modelname method significantly outperforms the other methods across all categories, particularly excelling in the 'Function' and 'Referring Expression' categories in both icon and text modalities. This suggests that \modelname's advanced algorithms are better at handling complex queries involving functional descriptions and referring expressions. On the other hand, methods such as R.W. with OCR and GPT-4o with G.P. show moderate performance, with GPT-4o with G.P. generally surpassing R.W. with OCR, especially in text-based processing.

\begin{table}[t]
\resizebox{\columnwidth}{!}{%
\begin{tabular}{cccccccccc}
\hline
\multirow{2}{*}{\textbf{Method}} &
  \multicolumn{2}{c}{\textbf{Name}} &
  \multicolumn{2}{c}{\textbf{Shape}} &
  \multicolumn{2}{c}{\textbf{Function}} &
  \multicolumn{2}{c}{\textbf{R.F.}} &
  \multicolumn{1}{l}{\multirow{2}{*}{\textbf{All}}} \\ \cmidrule(lr){2-3} \cmidrule(lr){4-5} \cmidrule(lr){6-7} \cmidrule(lr){8-9}
 &
  \textbf{Icon} &
  \textbf{Text} &
  \textbf{Icon} &
  \textbf{Text} &
  \textbf{Icon} &
  \textbf{Text} &
  \textbf{Icon} &
  \textbf{Text} &
  \multicolumn{1}{l}{} \\ \hline
R.W. w. OCR   & 0.09 & 0.12 & 0.04 & 0.09 & 0.07 & 0.13 & 0.15 & 0.22 & 0.13 \\
R.W. w. G.P. & 0.34 & 0.32 & 0.25 & 0.16 & 0.27 & 0.33 & 0.52 & 0.35 & 0.33 \\
GPT4o w. G.P.  & 0.27 & 0.16 & 0.19 & \textbf{0.27} & \textbf{0.44} & 0.31 & 0.40 & 0.33 & 0.28 \\
Auto-Explorer & \textbf{0.42} & \textbf{0.42} & \textbf{0.30} & 0.15 & 0.34 & \textbf{0.40}  & \textbf{0.67} & \textbf{0.50} & \textbf{0.42} \\ \hline
\end{tabular}%
}
\caption{\textbf{Accuracy Performance} (\%) over Different Query Types on\benchmark~benchmark.}
\label{tab:query}
\end{table}

\begin{figure*}[tbp]
    \centering
    \includegraphics[width=\textwidth]{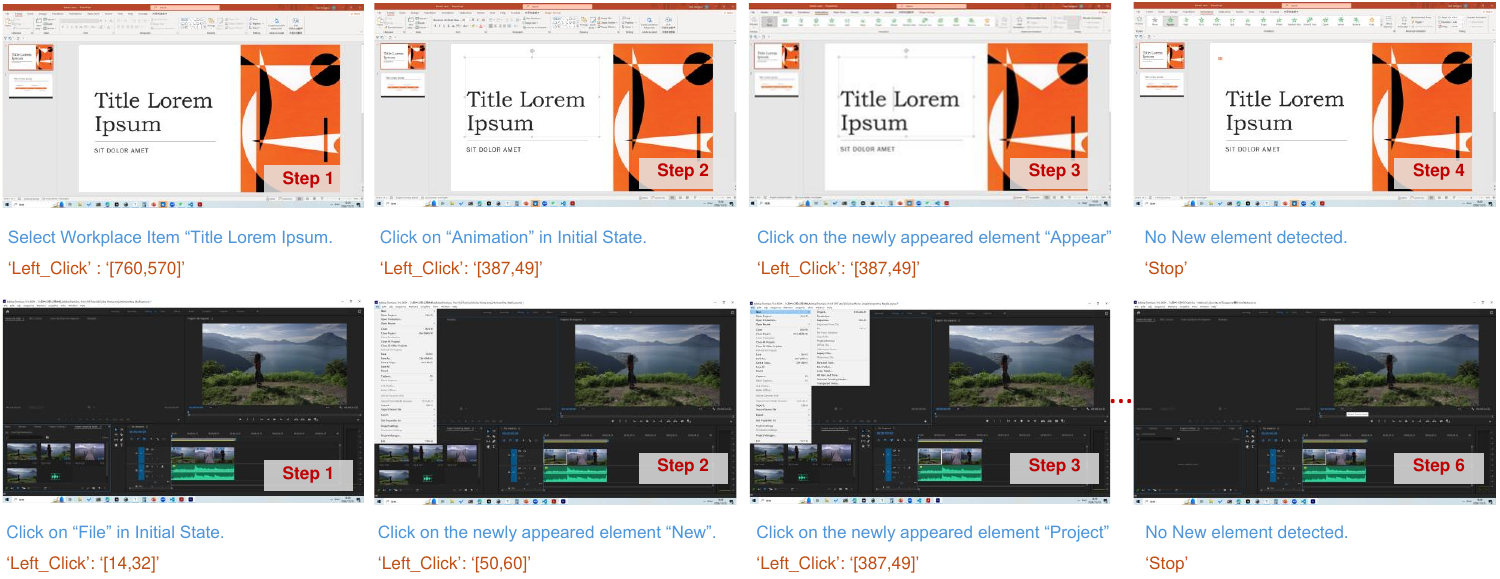}
    \caption{\textbf{Visualization of Some Exploration Action Trajectories.} \modelname~ not only discovers some different states of UI but also identifies meaningful trajectories.}
    \label{fig:trajectory}
\end{figure*}

\subsection{Performance by Data Volume}

The amount of training data typically has a significant impact on model performance. To investigate this issue, we experimented with allowing \modelname~to explore more or fewer steps on environments in\benchmark.
Figure~\ref{fig:data_volume} shows the relationship between the number of fine-tuning samples (in thousands) and the accuracy achieved by the category of the UI elements, as indicated by the line colors. As the number of fine-tuning samples increases from 2.5K to 17.5K, all methods show a clear upward trend in accuracy, demonstrating the effectiveness of additional training data in improving model performance. The Text Accuracy and Icon Accuracy lines are quite close throughout, suggesting a similar rate of improvement in both text and icon modalities. The All Accuracy line, representing the aggregate accuracy across both modalities, closely tracks the individual lines, indicating consistent performance improvements across different data types. Notably, with just a small number of training data, \modelname can surpass the SOTA method, OmniParser Accuracy. It suggests that data exploration itself is very helpful for supporting some new software.


\subsection{Impact of GUI-Parser on Experimental Results}
We designed two baselines, Random Walk, with different methods to capture GUI elements, using either OCR or our specially designed GUI parser, i.e., Random Walk w. OCR and Random Walk w. GUI Parser.
Table~\ref{tab:Table1} illustrates the performance of them. Notably, when Explorer combined with GUI Parser, significantly enhances performance across almost all tasks, particularly in software applications where both icon and text understanding are crucial. 

Moreover, it is important to note that even with OCR, performance on text is not very high despite the robustness of current OCR models. This is because the OCR parser tends to overlook many interactive buttons, preventing the exploration of additional states. This may be the primary reason for the poor performance of the Random Walk with OCR method.



\begin{table}[]
\resizebox{\columnwidth}{!}{%
\begin{tabular}{ccccccccc}
\hline
\multirow{3}{*}{\textbf{Method}} &
  \multicolumn{3}{c}{\textbf{Seen}} &
  \multicolumn{5}{c}{\textbf{Unseen}} \\
\cmidrule(lr){2-4} \cmidrule(lr){5-9}
 &
  \multicolumn{2}{c}{Software} &
  \multirow{2}{*}{All} &
  \multicolumn{2}{c}{Software} &
  \multicolumn{2}{c}{Web} &
  \multirow{2}{*}{All} \\
\cmidrule(lr){2-3} \cmidrule(lr){5-6} \cmidrule(lr){7-8}
                         & {\ul Icon} & {\ul Text} &      & {\ul Icon} & {\ul Text} & {\ul Icon} & {\ul Text} &      \\ \hline
R.W. w. OCR        & 0.11       & 0.22       & 0.17 & 0.04       & 0.12       & 0.13       & 0.20        & 0.11 \\
R.W. w. G.P. & 0.34       & \textbf{0.38}      & 0.36 & 0.22       & 0.20        & 0.51       & 0.37       & 0.20  \\
GPT4o w. G.P.      & 0.24       & 0.33       & 0.33 & 0.16       & 0.17       & 0.42       & 0.32       & 0.17 \\

Auto-Explorer            & \textbf{0.41}       & 0.38       & \textbf{0.40}  & \textbf{0.27}       & \textbf{0.31}       & \textbf{0.64}       & \textbf{0.49}       & \textbf{0.30}  \\ \hline
\end{tabular}%

}

\caption{\textbf{Accuracy Performance} (\%) over Seen and Unseen software or website domains on\benchmark~benchmark.}
\label{tab:seen}
\end{table}

\subsection{Visualization of Collected Action Trajectory}

In Figure~\ref{fig:trajectory}, we present two examples of the action trajectory of \modelname autonomously exploring environments. It is evident that \modelname not only effectively discovers different states but also discovers some meaningful trajectories. For instance, the example in PowerPoint involves modifying animations, and the Premiere Pro example explores a trajectory for creating a new project file. These data could also be helpful for the planning module in GUI Agent, which we will leave for future work. Additional trajectories explored by the model are showcased in the supplementary materials.

\section{Conclusion}

This paper explores the data collection challenges in the field of GUI agents, which enable the interpretation of natural language commands to operate software interfaces. Traditional data collection methods, such as HTML parsing, fall short in non-web environments due to the absence of easily accessible metadata. Addressing this, we introduce a novel GUI parsing model paired with an exploration mechanism to identify, localize, and interact with interface elements within various software applications. Our model utilizes Optical Character Recognition (OCR) and template-matching for element detection and includes a sophisticated exploration strategy that maximizes the observation of diverse UI elements. Furthermore, we establish a benchmark for evaluating the data exploration quality of GUI agents, focusing on their ability to gather diverse, meaningful data in productivity applications. Our approach demonstrates significant advantages in efficiency and coverage over existing methods, confirming its potential for GUI data collection.
{
\small
\bibliographystyle{ieeenat_fullname}
\bibliography{main}
}



\end{document}